\documentclass[10pt,twocolumn,letterpaper]{article}

\usepackage[pagenumbers]{cvpr} 
\usepackage{float}

\definecolor{cvprblue}{rgb}{0.21,0.49,0.74}
\usepackage[pagebackref,breaklinks,colorlinks,allcolors=cvprblue]{hyperref}

\def\paperID{*****} 
\def\confName{CVPR}
\def\confYear{2025}

\title{Optimizing YOLOv8 for Parking Space Detection: Comparative Analysis of Custom Backbone Architectures}

\author{
Apar Pokhrel \\
The University of Texas at Arlington \\
{\tt\small apar.pokhrel@mavs.uta.edu}
\and
Gia Dao \\
The University of Texas at Arlington \\
{\tt\small gia.daoduyduc@mavs.uta.edu}
}

\begin{document}
\maketitle
\begin{abstract}
Parking space occupancy detection is a critical component in the development of intelligent parking management systems. Traditional object detection approaches, such as YOLOv8, provide fast and accurate vehicle detection across parking lots but can struggle with borderline cases, such as partially visible vehicles, small vehicles (e.g., motorcycles), and poor lighting conditions. In this work, we perform a comprehensive comparative analysis of customized backbone architectures integrated with YOLOv8. Specifically, we evaluate various \textbf{backbones} -- ResNet-18, VGG16, EfficientNetV2, Ghost -- on the PKLot dataset in terms of detection accuracy and computational efficiency. Experimental results highlight each architecture's strengths and trade-offs, providing insight into selecting suitable models for parking occupancy.
\end{abstract}    
\section{Introduction}
\label{sec:intro}

Urban parking shortages and inefficient use of parking facilities are everyday challenges for drivers. Parking availability is a common challenge in urban environments, large
facilities, and university spaces. Traditional approaches have largely focused on using object detection algorithms, which aim to detect and localize vehicles in parking lot images or video feeds. Among these, YOLO (You Only Look Once)~\cite{10533619} has emerged as a popular real-time detection model, particularly with the introduction of YOLOv8.\\

Traditional methods relying solely on object detection often struggle with fine-grained classification (e.g., distinguishing between "occupied," "free," or "partially occupied" spots). However, in practical parking lot scenarios, vehicles may only partially occupy a space, non-vehicle objects may trigger false detections, and small vehicles like motorcycles can be easily missed. To address these limitations, this paper explores the impact of integrating different backbone architectures and head modifications into the YOLOv8 framework. By customizing the feature extraction backbone and detection head, we aim to enhance detection robustness and generalization. This comparative analysis provides valuable information on the strengths and trade-offs of each configuration, guiding the selection of optimal models for practical deployment in diverse parking scenarios.



\section{Related Work}
Different methods and techniques have been proposed to tackle the problem of detecting parking space occupancy. Nguyen and Sartipi introduced a novel automated parking space localization algorithm called \textbf{PakLoc}, complemented by \textbf{PakSke}, a module that refines the orientation and dimensions of the box bound. Furthermore, \textbf{PakSta}, an innovative framework that uses PakLoc's object detector that simultaneously monitors and detects the status of all parking spaces within a given frame, achieves an impressive AP75 of 93. 6\% in the PKLot dataset.~\cite{nguyen2024smartcameraparkingauto}.

Other detection methods used sensors for vehicle detection and surveillance. Boda and Howitt proposed design and implementation considerations for a wireless sensor network to track available spaces in public parking areas in real time. ~\cite {inproceedings}. The detector of the proposed system gave an output of 168 vehicles at the end of the simulation, while visually observing and recording a total of 171 vehicles during 40 minutes of data aggregation, which resulted in a probability of 1.8\% error.\\
Nguyen and Vo proposed a network based on the improved YOLOv5, named YOLO5PKLot, that focuses on redesigning the backbone network with a combination of the lightweight Ghost Bottleneck and Spatial Pyramid Pooling architectures. In addition, these researchers customized the head of YOLOv5 by resizing the anchors and adding one more layer of detection to improve the prediction task.~\cite{inproceedings_yolov5}. This method has significantly reduced the inference time and number of parameters used compared to YOLOv5m and YOLOv5s, with a 99.6\% detection accuracy. \\
The widespread adoption of Unmanned Aerial Vehicles (UAVs)  has introduced significant privacy challenges concerning unauthorized drone activities. To address these concerns, Wong et al we propose a novel model, WRN-YOLO\cite{Wong2025}, which integrates the Wide Residual Network (WRN) architecture with YOLO. 

\section{Dataset}

The project leverages the PKLot dataset~\cite{DEALMEIDA20154937}, which contains \texttt{12,417} images (\texttt{1280x720 pixels}) in JPEG color format, capturing various parking lots scenes, along with 695,899 segmented images of parking spaces. All images were acquired in the parking lots of the Federal University of Parana (UFPR) and the Pontifical Catholic University of Parana (PUCPR), located in Curitiba, Brazil. The images capture a variety of environmental conditions, including sunny, cloudy, and rainy weather, providing a diverse and realistic representation of parking scenarios. The images were taken over a period of more than 30 days with a 5-minute time-lapse interval using a low-cost full HD camera. The images were captured in two different parking lots, covering different weather conditions, such as sunny, cloudy, and rainy. The annotations of the original data set provided in XML format have been converted to YOLOv8 annotation format using Roboflow~\cite{dwyer2024roboflow}.\\
In this format, each image has a text file (.txt) where each line represents a normalized version of one parking space annotation --binary label indicating occupancy (0/1), horizontal center of the bounding box, vertical center of bounding box, width of the bounding box and the height of the bounding box - following the pattern: 
\[
\langle \text{class\_id},\ x_{\text{center}},\ y_{\text{center}},\ \text{width},\ \text{height} \rangle
\] 

\begin{figure}[h]
  \centering
   \includegraphics[width=0.75\linewidth]{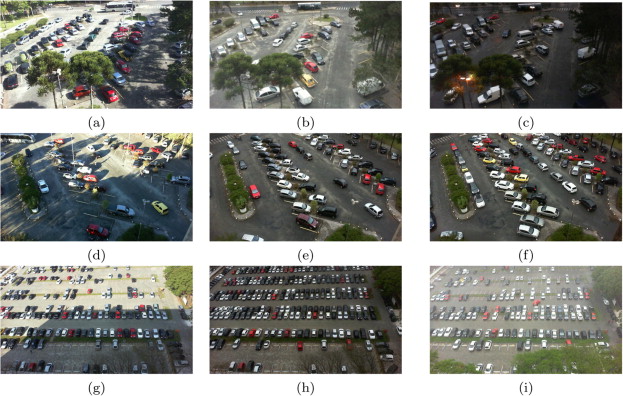}

   \caption{Figure 1: Images captured under different weather conditions: (a) sunny (b) overcast, and (c) rainy
from UFPR04; (d) sunny (e) overcast, and (f) rainy from UFPR05; and (g) sunny (h) overcast, and (i) rainy from PUCPR}
\end{figure}

\begin{figure}[h]
  \centering
   \includegraphics[width=0.75\linewidth]{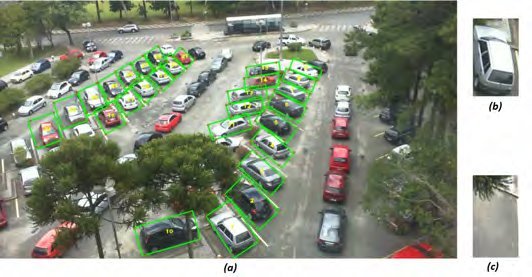}

   \caption{Segmented image: (a) 28 delimited spaces, (b) occupied sub-image, and (c) empty sub-image.}
\end{figure}

\section{YOLOv8 Architecture}
\begin{figure}[h]
  \centering
   \includegraphics[width=0.75\linewidth]{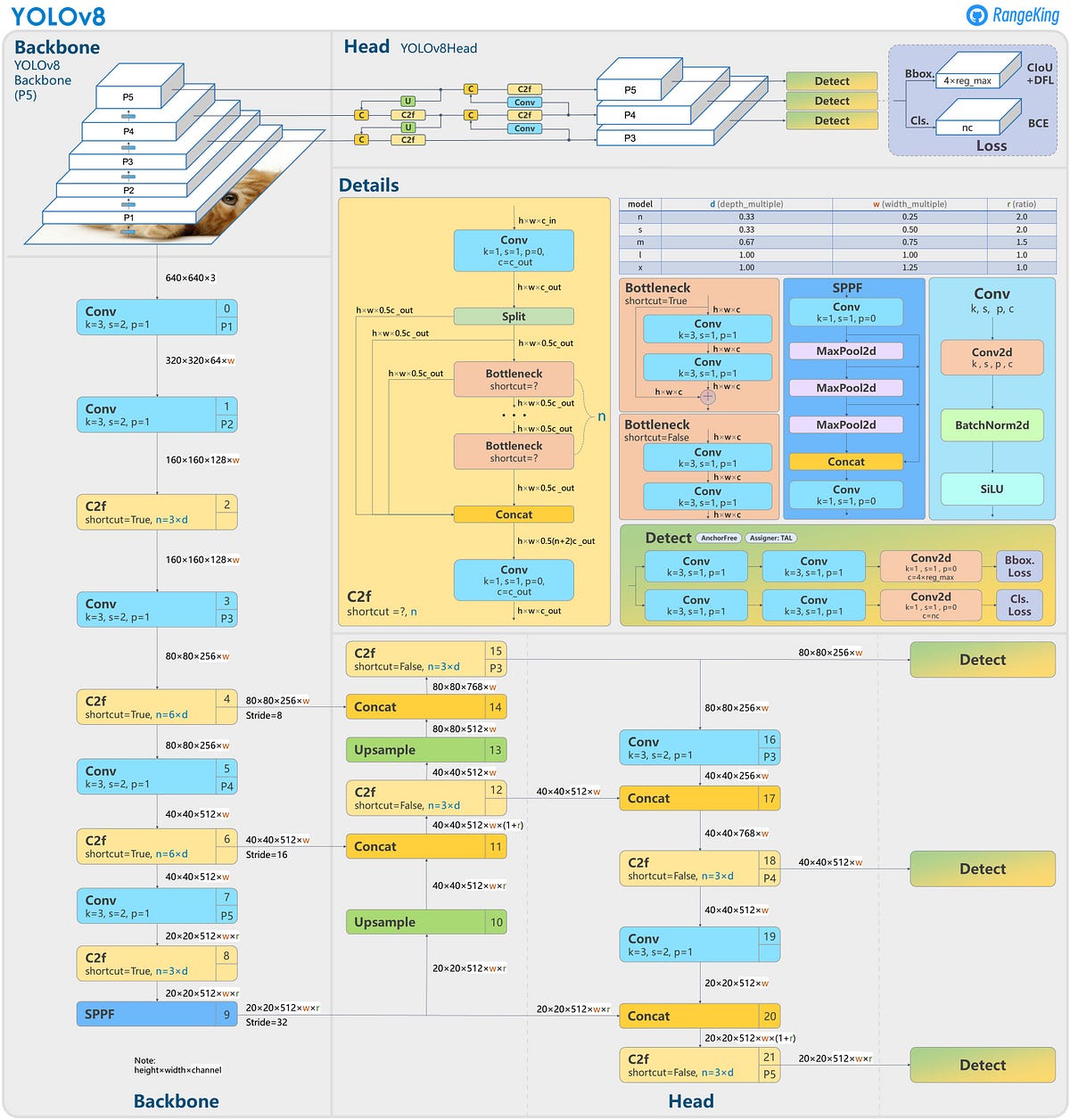}
   \caption{An overview of YOLOv8 architecture}
   \label{fig:yolov8_architecture}
\end{figure}

YOLOv8~\cite{yolov8} is the latest iteration of the You Only Look Once (YOLO) family of object detection models. It was developed and released by the Ultralytics team in January 2023. Traditional methods relied heavily on sliding-window approaches. YOLO treats object detection as a single regression problem by simultaneously predicting all bounding boxes in a single network pass. YOLOv8 improves on its predecessor YOLOv5~\cite{yolov5} by utilizing anchor-free detection, which expedites nonmaximum suppression (NMS) post-processing, a step that discards overlapping boxes with lower confidence ratings. It also employs state-of-the-art backbone and neck architectures, resulting in improved feature extraction and object detection performance. The YOLOv8 series has a diverse range of models catered towards object detection, instance segmentation, image classification, pose estimation, and multi-object tracking. It also offers various sizes, from nano-sized (n) to x-large (x). Figure~\ref{fig:yolov8_architecture} by GitHub user RangeKing shows the detailed visualization of the network's architecture. Ultralytics explicitly does not label these parts in their official documentation, but the division is commonly accepted in the community.

    \subsubsection{Backbone}
    The backbone is responsible for extracting features from input images. YOLOv8 uses a custom version of the CSPDarknet53 ~\cite{bochkovskiy2020yolov4optimalspeedaccuracy}, which uses Cross-Stage Partial (CSP) ~\cite{wang2019cspnetnewbackboneenhance} connections to improve information flow between layers. CSP divides the input feature map into two parts: one is processed through a series of bottleneck layers to capture deep semantic features, while the other is forwarded directly to preserve low-level spatial information. This design reduces computational complexity, enhances gradient flow while reducing model size.

    \subsubsection{Neck}
    The neck serves as a bridge between the backbone and the head. The neck is responsible for fusing the features extracted by the backbone. SPFF (Spatial Pyramid Pooling) block at the neck provides a multi-scale representation of the feature map. It employs a PANet (Path Aggregation Network)~\cite{liu2018pathaggregationnetworkinstance}, which combines features at different scales from different stages of the backbone. YOLOv8 utilizes a novel C2f module instead of the traditional Feature Pyramid Network (FPN). The C2f module helps in efficient feature map refinement and helps reduce memory consumption. The Convolutional layers \textbf{P3}, \textbf{P4}, and \textbf{P5} are transmitted to various parts of the pyramid (layers 11, 14, and 20), which allows the model to detect objects of various sizes.
    
    \subsubsection{Head}
    The detection head uses Dynamic Anchor Boxes and a novel IOU loss function. The head consists of three detection heads connected to the three outputs of the PANet. ~\cite{yaseen2024yolov8indepthexplorationinternal} These heads are responsible for generating bounding boxes, assigning confidence scores, and classifying boxes into their categories. The first Detect block specializes in detecting small objects from the C2f block present in Block 15. The second Detect block specializes in detecting medium objects from the C2f block present in Block 18. Lastly, the third Detect block specializes in detecting small objects from the C2f block in Block 21.

\section{Methodology}
\label{sec:formatting}

This study investigates the effectiveness of different backbone architectures in enhancing parking space occupancy detection using YOLOv8. The overall process involves integrating alternative backbones into the YOLOv8 framework to assess their impact on detection performance. The chosen backbones -- ResNet18, EfficientNetV2, Ghost, and VGG16 -- offer diverse trade-offs in terms of depth, parameter efficiency, and representational power.
\subsection{Dataset Preparation}
YOLOv8 uses the YOLOv5 PyTorch TXT annotation format, a modified version of the Darknet annotation format. The original images are resized to a dimension \texttt{640x640} pixels. The dataset is divided into 70\% for training, 20\% for validation, and 10\% for testing. During training, data was augmented by applying random horizontal flipping, brightness, HSV augmentation, image translation, image scaling, blur, and some variation of mosaic augmentation from YOLO. 

\subsection{YAML Configuration}
YOLOv8 allows model customization via a YAML configuration file. Ain't Markup Language (YAML) is a human-readable data serialization language used for storing and organizing data, especially configuration settings. The YAML file for YOLOv8 defines the model architecture by specifying the layers and their configurations. When loading or training a model, the YAML file is passed to a parser that dynamically constructs the model's architecture at runtime. The YAML configuration for YOLOv8 typically consists of: number of classes (nc), scales, and the layer definitions grouped under backbone and head. Each entry in the layer is represented as a list in the form 
\texttt{[from, number, module, args]}.
Each custom backbone is implemented as a torchvision model and plugged into the YAML file.

\begin{figure}[h]
  \centering
   \includegraphics[width=0.75\linewidth]{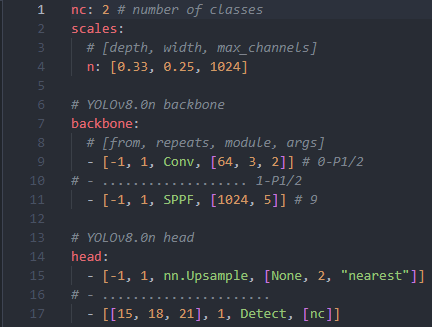}
   \caption{A sample YAML configuration of YOLOv8n}
    \label{fig:yaml}
\end{figure}

\subsection{YOLOv8n}
We focus mainly on object detection with a scale variant of nano (n). We used pre-trained weights of the YOLO8n from the official repository to load the model, and fine-tuned the model on the PkLot dataset. YOLO8n is treated as the base model. 
    
\subsection{YOLO-ResNet-18}
\begin{figure}[h]
  \centering
   \includegraphics[width=0.75\linewidth]{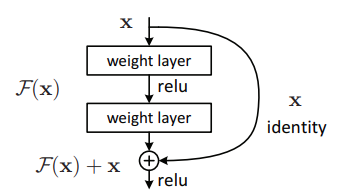}
   \caption{Residual Block in ResNet18}
    \label{fig:resnet_block}
\end{figure}

ResNet-18 is a convolutional neural network from the ResNet~\cite{he2015deepresiduallearningimage} family. ResNet-18 is one of the shallower versions of ResNet, consisting of 18 layers, including convolutional layers and residual blocks. A residual block allows inputs to bypass one or more layers via shortcut connections, called skip connections.  Residual connections help avoid vanishing gradients for deeper networks. A typical residual block, as shown in Figure~\ref{fig:resnet_block}, consists of two convolutional layers with batch normalization, ReLU activation, and a skip connection. ResNet18 consists of an initial \texttt{7X7} convolution layer followed by a MaxPooling layer. This is followed by four stages of residual blocks. Finally, the output is passed through an average pooling layer followed by a fully connected layer for the final classification output.

We replaced the default YOLOv8 backbone with \texttt{ResNet18} having pre-trained weights from the torchvision.models library. A custom YAML file \texttt{yolov8n-resnet18.yaml} with ResNet18 configuration was created. We extracted feature maps from intermediate layers - Layers \textbf{6}, \textbf{7}, and \textbf{8} of ResNet18- that correspond to spatial resolutions of 1/8 (P3), 1/16 (P4), and 1/32 (P5) of the input image. These layers were selected using the Index module in YOLO’s config to map to internal blocks of ResNet18. An SPPF (Spatial Pyramid Pooling – Fast) module was applied to the deepest output to enhance the receptive field and capture multi-scale context. The final detection head makes predictions from fused feature maps of Layers \textbf{10}, \textbf{13}, and \textbf{16}. A YOLO was then initialized with the YAML file and trained.

\subsection{YOLO-EfficientNetV2}
EfficientNetV2~\cite{tan2021efficientnetv2smallermodelsfaster} is part of the EfficientNet family of convolutional neural networks introduced by Tan and Le, designed to optimize both accuracy and training speed. Unlike traditional architectures, EfficientNetV2 combines MBConv and Fused-MBConv blocks to improve parameter efficiency, convergence speed, and representational power. These blocks use depthwise separable convolutions, squeeze-and-excitation (SE) modules, and shortcut connections to reduce computation while retaining expressive capacity. A typical EfficientNetV2 architecture begins with a stem convolution followed by a series of progressively deeper stages, each composed of multiple MBConv or Fused-MBConv blocks. These stages extract features at increasing levels of abstraction. Unlike ResNet, which uses uniform residual blocks, EfficientNetV2 dynamically chooses block types and expansion ratios per stage, guided by neural architecture search.

We replaced the default YOLOv8 backbone with \texttt{EfficientNetV2-S}, having pre-trained weights from the torchvision.models library. A custom YAML file \texttt{yolov8n-efficientv2.yaml} with EfficientNetV2-S configuration was created. We extracted feature maps from intermediate layers - Layers \textbf{4}, \textbf{6}, and \textbf{8} of EfficientNetV2-S- that correspond to spatial resolutions of 1/8 (P3), 1/16 (P4), and 1/32 (P5) of the input image. These layers were selected using the Index module in YOLO’s config to map to internal blocks of EfficientNetV2. An SPPF  module was applied to the deepest output to enhance the receptive field and capture multi-scale context. The neck and head were modified to match the feature shapes of the EfficientNetV2 outputs. As in YOLO-ResNet18, the final detection head takes fused outputs from the PANet-style neck at three scales from Layers \textbf{10}, \textbf{13}, and \textbf{16} and predicts bounding boxes, objectness scores, and class labels. A YOLO model was then initialized with the YAML file and trained.

\begin{figure}[h]
  \centering
   \includegraphics[scale=0.6]{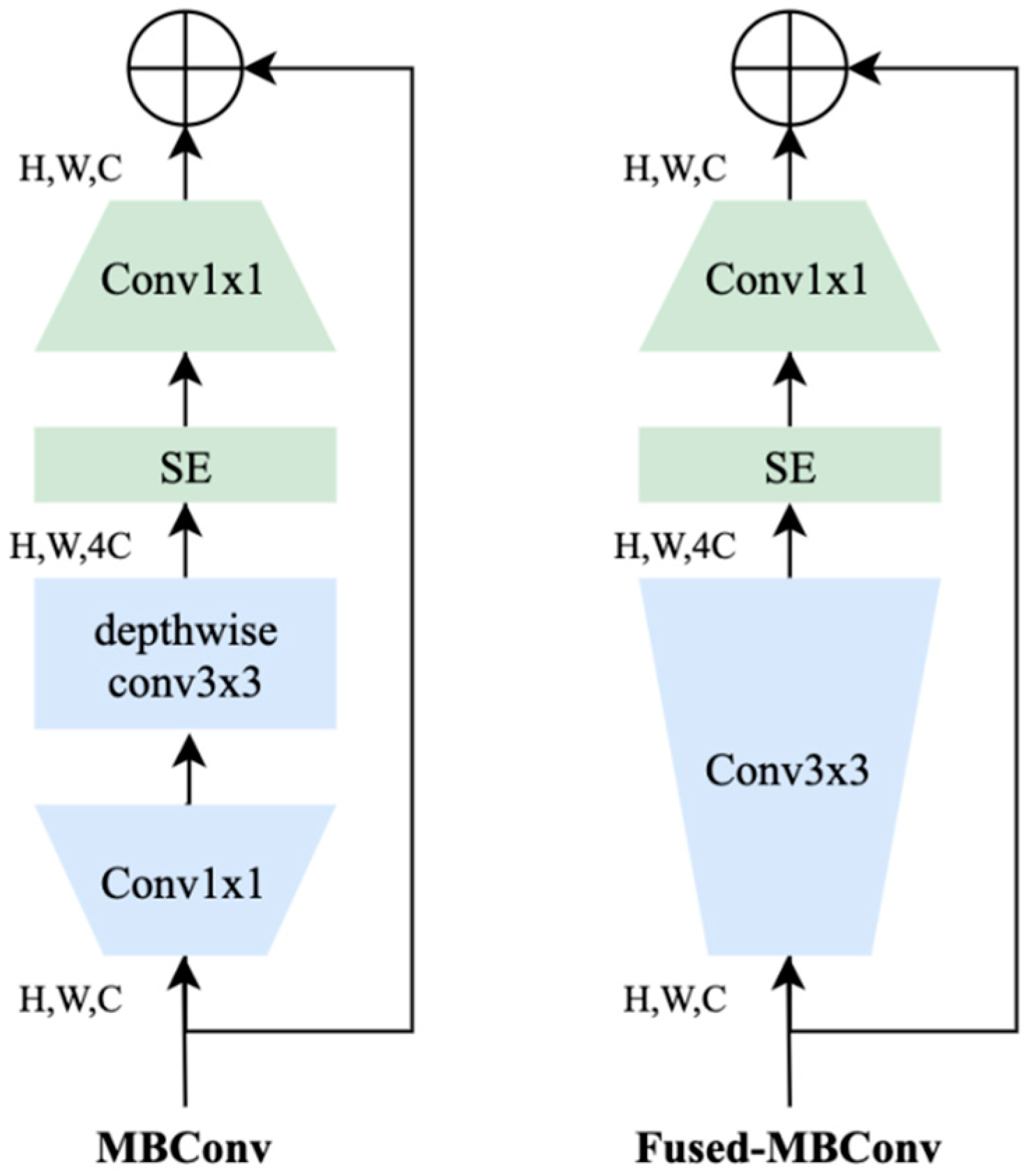}
   \caption{Structure of MBConv and Fused-MBConv}
   \label{fig:efficient_architecture}
\end{figure}

\subsection{YOLO-VGG-16}
The \texttt{VGG-16} architecture, proposed by Simonyan and Zisserman~\cite{simonyan2015deepconvolutionalnetworkslargescale}, represents a significant improvement over earlier convolutional neural networks (ConvNets) by increasing network depth through the addition of multiple convolutional layers, all employing small 3x3 kernel filters.

The \texttt{VGG-16} architecture is characterized by its simplicity, comprising a stack of convolutional layers followed by max-pooling layers. Each convolutional layer utilizes a very small 3x3 receptive field with 1-pixel spatial padding to preserve spatial dimensions after performing convolution. Five max-pooling layers, applied after selected convolutional layers, perform spatial downsampling over a 2x2 window with a stride of 2. The use of 3x3 kernels significantly reduces the number of parameters compared to larger kernels (e.g., 7x7), while incorporating multiple non-linear rectification layers (ReLU) enhances the discriminative power of the decision function compared to architectures with fewer non-linearities.

To enhance the feature extraction capabilities of YOLOv8 for object detection, we replaced its default backbone with the \texttt{VGG-16} architecture, while utilizing the \texttt{VGG-16} pretrained weights from the torchvision.models library. A custom YAML file \texttt{yolov8n-vgg16.yaml} was created to integrate specific VGG-16 layers into the YOLOv8 framework, replacing the original backbone. Specifically, we extracted feature maps from \texttt{VGG-16} layers \textbf{23}, \textbf{30}, and \textbf{31}, corresponding to spatial resolutions of 1/8 (P3), 1/16 (P4), and 1/32 (P5). These layers were selected using the Index module in YOLO’s config to map to internal blocks of \texttt{VGG-16}. An SPPF module was applied to the deepest output to enhance the receptive field and capture multi-scale context. The final detection head takes fused outputs from the PANet-style neck at three scales from Layers \textbf{10}, \textbf{13}, and \textbf{16} and predicts bounding boxes, objectness scores, and class labels. A YOLO model was then initialized with the custom YAML file and trained.


\subsection{YOLO-Ghost-P2}
YOLOv8-Ghost-P2 is a lightweight variant of the YOLOv8 model family. Specifically, Ghost-P2 utilizes GhostConv layers to replace the original Conv, which significantly reduces the number of parameters to compute while maintaining an efficient way of generating more features. Ghost Module, originally proposed by Noah's Ark Lab and Huawei Technologies, tackles the issue of high parameters and FLOPS induced by normal convolution. 
Initially, the input tensors go through the ordinary convolutional operation to obtain a smaller size of intrinsic feature maps \( Y' \). 

Specifically, \(m\) intrinsic feature maps \( Y' \in \mathbb{R}^{h' \times w' \times m} \) are generated using a primary convolution:
\[Y' = X * f'\]
where \( f' \in \mathbb{R}^{c \times k \times k \times m} \) are the learned filters, \( m' \) is the number of output channels, \(m < n\) and the bias term is omitted for simplicity.
A cheap linear operation \( \Phi \) is applied on these intrinsic feature maps to produce an additional ghost (similar pairs) feature maps.~\cite{han2020ghostnetfeaturescheapoperations}. The final output of the Ghost Module stacks the intrinsic feature maps with their corresponding ghosts, producing a total of \(n = m * s\) feature maps, where \(s\) is the number of ghost feature maps generated for each original feature. This approach ensures that the output maintains the necessary spatial dimensions while reducing computational complexity.

\begin{figure}[h]
    \includegraphics[scale=2.4]{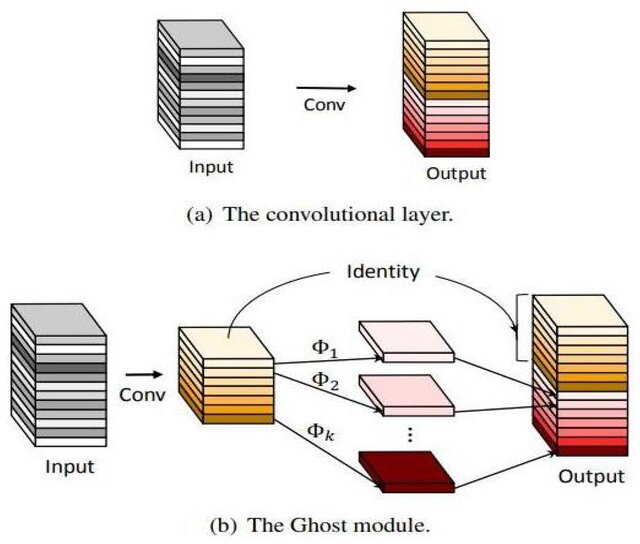}
    \caption{An illustration of a convolutional layer and the Ghost module operation for outputting the same number of feature maps}
\end{figure}

\section{Evaluation  Metrics}
\label{sec:metrics}
To evaluate the effectiveness of our models, we employ distinct performance metrics.

\subsection{Intersection of Union (IOU)}
\begin{figure}[h]
  \centering
   \includegraphics[width=0.75\linewidth]{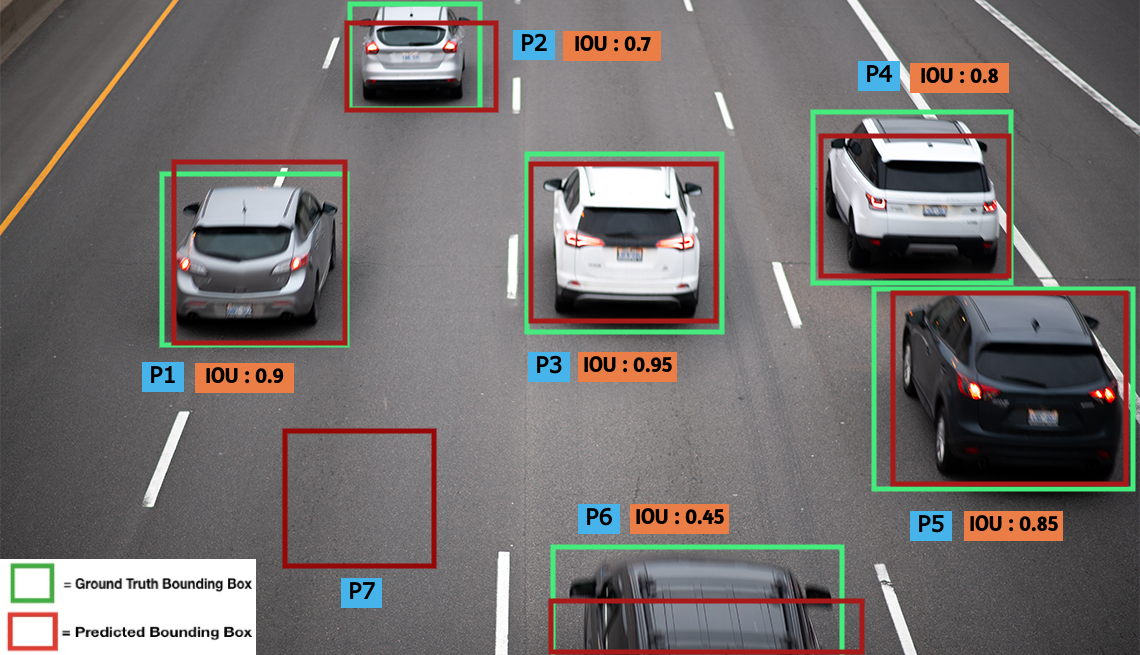}

   \caption{Truth and Predicted Bounding Box}
   \label{fig:bounding boxes}
\end{figure}

IOU is a metric that quantifies the degree of overlap between two regions, as shown in Figure~\ref{fig:bounding boxes}. The value ranges from 0 to 1. IoU evaluates the localization accuracy of detected bounding boxes by measuring their overlap with ground-truth annotations. It is defined as: \\
\begin{equation}
    IoU = \frac{|A \cap B|}{|A \cup B|}
\end{equation}

\subsection{Mean Average Precision}
To evaluate detection performance, we adopt the standard \textit{mean Average Precision} (mAP) metrics, which are based on the Precision-Recall (PR) curve. For a given class \( c \), the \textit{Average Precision} (AP) is defined as the area under the PR curve:

\begin{equation}
\text{AP}_c = \int_0^1 p_c(r) \, dr
\end{equation}

where \( p_c(r) \) denotes the precision as a function of recall \( r \). In practical implementations, this integral is approximated over discrete recall thresholds.

The \textit{mean Average Precision} (mAP) across all \( C \) classes is computed as:

\begin{equation}
\text{mAP} = \frac{1}{C} \sum_{c=1}^{C} \text{AP}_c
\end{equation}

For localization accuracy, AP is typically calculated at various Intersection-over-Union (IoU) thresholds.

The metric \textbf{mAP@50} corresponds to an IoU threshold of 0.5. To provide a more comprehensive evaluation, the \textbf{mAP@50:95} metric averages AP over 10 IoU thresholds from 0.50 to 0.95 with a step size of 0.05. This gives a more rigorous evaluation across a range of IoU thresholds.

\begin{table}
    \centering
    \begin{tabular}[h]{|p{1cm}|p{5.4cm}|}
        \hline
        \textbf{Term} & \textbf{Description} \\ \hline
        TP & Correctly identified occupied spaces \\ \hline
        TN & Correctly identified empty spaces \\ \hline
        FP & predicted an empty space as occupied \\ \hline
        FN & predicted an occupied space as empty \\ \hline
    \end{tabular}
   \caption{Prediction Outcomes.}
\label{tab:metrics_definitions}
\end{table}

\begin{figure}[h]
  \centering
   \includegraphics[width=0.75\linewidth]{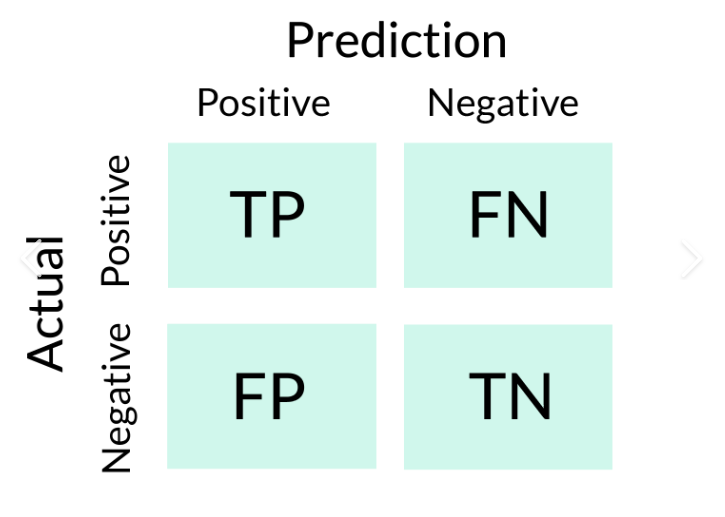}
   \caption{Actual vs Prediction}
   \label{fig:ac-vs-pred}
\end{figure}

\subsection{Precision}
It measures the proportion of correctly predicted occupied spaces out of all spaces predicted as occupied. A higher precision value indicates fewer false positives.  FP, FN, TP, and TN stand for False Positive, False Negative, True Positive, and True Negative
Negative, respectively. These statistics are defined in the 2 X 2 confusion matrix as shown in Figure~\ref{fig:ac-vs-pred} and explained in Table~\ref{tab:metrics_definitions}.

\begin{equation}
    Precision = \frac{TP}{TP + FP}
\end{equation}

\subsection{Recall}
 It measures the proportion of correctly predicted occupied spaces out of all actual occupied spaces. A higher recall indicates fewer false negatives.
\begin{equation}
    Precision = \frac{TP}{TP + FN}
\end{equation}

\section{Experiment and Results}

\subsection{Experimental Setup}
All experiments were conducted using an NVIDIA A100-SXM4-40GB GPU hosted in a CUDA 12.4 environment with driver version 550.54.15 and Pytorch 2.6. Ultralytics YOLOv8 was used for all YOLO models. The GPU offered ample compute capability (sm\_80) and 40 GB of VRAM, which allowed for efficient training of multiple YOLOv8 variants with large batch sizes and high-resolution inputs. Some experiments were conducted using an NVIDIA Tesla T4 GPU to reduce GPU compute cost. 

All YOLO variants were trained and evaluated under identical conditions to allow for fair comparison. Table~\ref{tab:training-params} highlights the parameters used during training YOLO variants on the PkLot dataset. Empty slots are labeled as \textbf{e} and occupied slots are labeled as \textbf{o} sometimes.

\begin{table}[t]
    \centering
    \caption{Training configuration and hyperparameters}
    \label{tab:training-params}
    \begin{tabular}{@{}ll@{}}
    \toprule
    \textbf{Parameter} & \textbf{Value} \\
    \midrule
    Input Size        & 640 $\times$ 640 \\
    Epochs                  & 20  \\
    Optimizer               & AdamW (auto-selected) \\
    Initial Learning Rate   & 0.001667 (auto-tuned) \\
    Momentum                & 0.9 (auto-tuned) \\
    Weight Decay            & 0.0005 (for conv weights) \\
    Bias Decay              & 0.0 \\
    Other Weight Decay      & 0.0 \\
    Batch Size              & 16 \\
    Dataloader Workers      & 2 \\
    Logging Directory       & \texttt{runs/detect/train} \\
    \bottomrule
    \end{tabular}
\end{table}

\subsection{Results}

All YOLO variants were evaluated on the PkLot dataset using the metrics described in \hyperref[sec:metrics]{Section 6}. The standard metrics are shown in Table~\ref{tab:training-params}. Additionally, we report model complexity in terms of parameter count, number of layers, and inference speed, which are shown in Table~\ref{tab:extended_comparison}.

All models achieved high precision and recall, but subtle differences emerge when comparing stricter metrics like mAP50:95. \textbf{YOLO-EfficientNetv2} achieved the best overall detection performance, with the highest mAP50:95 (\texttt{0.986}) and excellent precision (\texttt{0.998}), while maintaining a moderate computational cost (\texttt{56.4} GFLOPs) and lower inference time (\texttt{4.1} ms).
YOLO-ResNet-18 offered a strong balance, matching the mAP50 score (\texttt{0.994}) and achieving strong precision (\texttt{0.998}) with slightly lower localization accuracy (mAP50:95 = \texttt{0.976}).

YOLO-VGG16, while achieving high precision (\texttt{0.998}), showed a slightly lower recall (\texttt{0.985}) and mAP50:95 (\texttt{0.985}), and was the most computationally expensive model (\texttt{262.1} GFLOPs), despite a moderate inference time (\texttt{3.3} ms), making it less suitable for lightweight deployment. YOLOv8n maintained competitive precision and recall (both \texttt{0.996}) and fast inference (\texttt{0.9 ms}), demonstrating its suitability for real-time applications with minimal compute.
YOLO-Ghost-P2 achieved the fastest inference (\texttt{1.5 ms}) among custom modifications with the lowest parameter count (\texttt{1.6M}), but its slightly lower recall (\texttt{0.978}) and mAP50:95 (\texttt{0.896}) suggest a higher likelihood of missed detections.

Overall, models like YOLO-VGG16 and EfficientNet excel in detection accuracy, while YOLOv8n and Ghost-P2 are better suited for edge applications where inference speed and model size are critical. Precision-recall trade-offs highlight the need to match model choice with the deployment scenario’s tolerance for false positives and missed detections.

\begin{table}[t]
\centering
\small
\setlength{\tabcolsep}{5pt} 
\caption{Comparison of YOLO Variants on PKLot Dataset}
\begin{tabular}{p{2.4cm}p{1.2cm}ccc c}
\toprule
\textbf{Model} & \textbf{Precision} & \textbf{Recall} & \textbf{mAP50} & \textbf{mAP50:95} \\
\midrule
YOLOv8n & 0.996 & 0.996 & 0.994 & 0.97 \\
YOLO-ResNet-18  & 0.998 & 0.997 & 0.994 & 0.976\\
YOLO-VGG16 & 0.998 & 0.985 & 0.991 & \textbf{0.985} \\
\midrule
YOLO-EfficientNet  & 0.998 & 0.997 & 0.994 & \textbf{0.986} \\
YOLO-Ghost-P2  & 0.968 & 0.978 & 0.991 & 0.896 \\

\bottomrule
\end{tabular}
\label{tab:yolo_comparison}
\end{table}

\begin{table}[t]
\centering
\small
\setlength{\tabcolsep}{4pt}
\caption{Extended Comparison of YOLO Variants}
\begin{tabular}{p{2.4cm}p{1.2cm}p{1.2cm}@{\hskip 10pt}p{1.2cm}p{1.2cm}}
\toprule
\textbf{Model} & \textbf{Params (M)} & \textbf{Layers} & \textbf{Inference (ms)} & \textbf{GFLOPs} \\
\midrule
YOLOv8n            & 3.01 & 129 & \textbf{0.9} & 8.2 \\
YOLO-ResNet-18     & 13.32 & 132 & 9 & 35.2 \\
YOLO-VGG16         & 17.78 & 113 & 3.3 & 262.1 \\
\midrule
YOLO-EfficientNet  & 23.40 & 564 & 4.1 & 56.4 \\
YOLO-Ghost-P2      & 1.60 & 290 & \textbf{1.5} & 8.8 \\
\bottomrule
\end{tabular}
\label{tab:extended_comparison}
\end{table}

\subsection{Visualization}

\subsubsection{Plots}

\begin{figure}[h]
  \centering
   \includegraphics[width=0.75\linewidth]{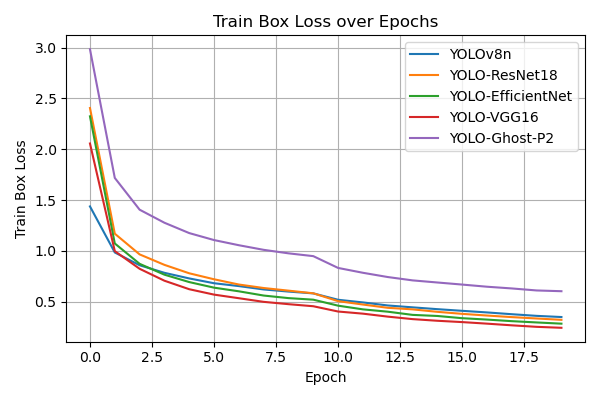}
   \caption{Train Box Loss- YOLO variants}
   \label{fig:com_train_box_loss]}
\end{figure}

\begin{figure}[h]
  \centering
   \includegraphics[width=0.75\linewidth]{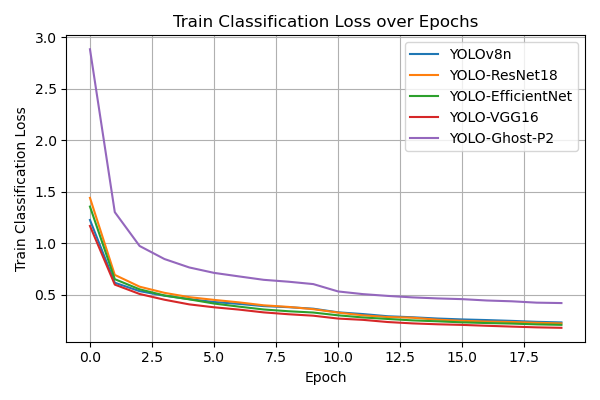}
   \caption{Train Classification Loss- YOLO variants}
   \label{fig:comp_train_class_loss]}
\end{figure}

\begin{figure}[h]
  \centering
   \includegraphics[width=0.75\linewidth]{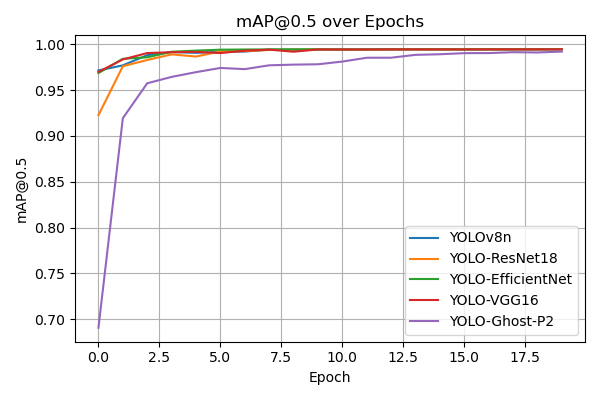}
   \caption{mAp@50- YOLO variants}
   \label{fig:comp_mAP50]}
\end{figure}

To better understand the training behavior of each YOLO variant, we present key performance metrics plotted over training epochs. Figures ~\ref{fig:com_train_box_loss]}, ~\ref{fig:comp_train_class_loss]}, ~\ref{fig:comp_mAP50]}, ~\ref{fig:comp_mAP50:95} show comparisons of each YOLO variants under box loss, class loss, mAP50, and mAP50:95. YOLOv8n and YOLO-Ghost-P2 showed the fastest drop in loss, attributed to their lightweight architectures, while YOLO-EfficientNet and YOLO-VGG16 required more epochs to stabilize due to deeper networks. Other plots are readily available in the code repository.

\begin{figure}[h]
  \centering
   \includegraphics[width=0.75\linewidth]{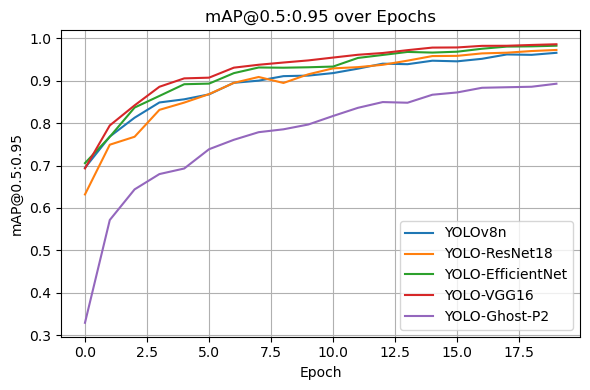}
   \caption{mAp@50:95- YOLO variants}
   \label{fig:comp_mAP50:95}
\end{figure}

\begin{figure}[h]
  \centering
   \includegraphics[width=0.75\linewidth]{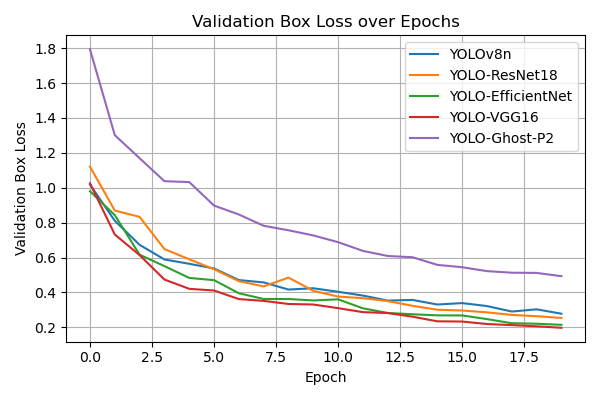}
   \caption{Validation Box Loss - YOLO variants}
   \label{fig:comp_mAP50:95}
\end{figure}

\section{Conclusion}

In this work, we investigated the impact of integrating custom backbone architectures into the YOLOv8 framework using the PKLot dataset. We evaluated five models—YOLOv8n, YOLO-ResNet18, YOLO-VGG16, YOLO-EfficientNet, and YOLO-Ghost-P2—across accuracy and efficiency metrics. Our results show that YOLO-EfficientNet achieved the best overall detection performance, combining high precision, recall, and localization accuracy with reasonable computational cost. These findings highlight the importance of backbone selection in adapting object detection models to specific application constraints. All code, training logs, and model configurations and weights are available at: \href{https://github.com/pokhrelapar/yolov8-pklot}{YOLOv8-PkLot}
\section{Challenges}
We faced some significant challenges throughout the project. Initial efforts of local installation of PyTorch Cuda, NVIDIA CUDA Toolkit, and cuDNN slowed down our experimental setup. As a result, we switched over to a cloud-hosted compute, Google Colab Pro environment for our needs. Customizing YOLOv8 with different
backbones significantly increases the computational demand for model training. This also required manual adjustments to feature map shapes and channel dimensions to ensure compatibility with the neck and head. Due to limited compute resources, we also fell behind on proper model configurations and hyperparameter tuning. The PkLot dataset, while diverse, includes fixed camera angles and limited parking lot types.
\section{Future Work}
To build upon our current YOLOv8 customization efforts, future work will involve exploring deeper backbone architectures, such as ResNet50, Vision Transformers (ViT), and Swin Transformer, to further improve detection accuracy. Incorporating additional modalities such as temporal data (video sequences), depth information could improve robustness. The integration of MobileNet architectures could also be helpful when deploying to edge devices for real-time detection on embedded systems (e.g. Jetson Nano or Xavier). 
\section{Acknowledgments}

We would like to thank the open-source community for their existing work on custom YOLO modifications across various domains. We also extend our sincere gratitude to Dr. Diego Patiño, Dr. Alex Dillhoff, and a few CSE doctoral students for their guidance and feedback, which helped shape the direction and depth of this project.

{
    \small
    \bibliographystyle{ieeenat_fullname}

}


\end{document}